%% file: root.tex
\documentclass{ifacconf}
\usepackage{graphicx}
\graphicspath{{figure/}{graphics/plots/}}
\usepackage[width=1\textwidth,format=hang]{caption}
\usepackage{bm}

\usepackage{natbib}
\usepackage{tikz}
\usepackage{adjustbox}
\usepackage{pgfplots}

\makeatletter
\let\old@ssect\@ssect 
\makeatother

\usepackage[hidelinks]{hyperref}
\makeatletter
\def\@ssect#1#2#3#4#5#6{%
  \NR@gettitle{#6}
  \old@ssect{#1}{#2}{#3}{#4}{#5}{#6}
}
\makeatother

\usepgfplotslibrary{groupplots}
\usetikzlibrary{pgfplots.groupplots}
\usetikzlibrary{plotmarks}
\usetikzlibrary{calc}
\usepackage{comment}
\usepgfplotslibrary{external}
\pgfplotsset{compat=newest}

\usetikzlibrary{matrix}
\pgfplotsset{grid style={dashed,gray}}
\usepackage{amsmath} 
\usepackage{amssymb}  
\usepackage{color,soul}
\usepackage{bibrange}
\usepackage{hvline}
\usepackage{url}
\usepackage{multirow}
\usepackage{booktabs}
\usepackage{epstopdf}
\usepackage{subfig}

\usepackage[ruled,linesnumbered,algo2e]{algorithm2e}
\usepackage{parskip}
\usepackage{hyphenat}
\usepackage{bm}
\makeatletter
\usepackage{siunitx}
\usepackage{nicematrix}
\newcommand{\nosemic}{\renewcommand{\@endalgocfline}{\relax}}
\newcommand{\dosemic}{\renewcommand{\@endalgocfline}{\algocf@endline}}

\providecommand{\Comma}{\text{~,\xspace}}

\begin{document}
\begin{frontmatter}
\title{Two-Step Online Trajectory Planning of a Quadcopter in Indoor Environments  with Obstacles}
\author[First]{M. Zimmermann},
\author[First]{M.N. Vu},
\author[First]{F. Beck},
\author[Third]{A. Nguyen},
\author[First,Second]{A. Kugi}

\address[First]{Automation and Control Institute (ACIN), 
   TU Wien, Vienna, Austria (e-mail: martinzimmermann1@gmx.at, \{vu,beck, kugi\}@acin.tuwien.ac.at).}
\address[Second]{Center for Vision, Automation and Control, AIT Austrian Institute of Technology, Vienna, Austria (e-mail: \{andreas.kugi\}@ait.ac.at)}
\address[Third]{Department of Computer Science, University of Liverpool, Liverpool, England (e-mail: anh.nguyen@liverpool.ac.uk)}

\begin{abstract}                
This paper presents a two-step algorithm for online trajectory planning in indoor environments with unknown obstacles. 
In the first step, sampling-based path planning techniques such as the optimal Rapidly exploring Random Tree (RRT*) algorithm and the Line-of-Sight (LOS) algorithm are employed to generate a collision-free path consisting of multiple waypoints. 
Then, in the second step, constrained quadratic programming is utilized to compute a smooth trajectory that passes through all computed waypoints. 
The main contribution of this work is the development of a flexible trajectory planning framework that can detect changes in the environment, such as new obstacles, and compute alternative trajectories in real time. 
The proposed algorithm actively considers all changes in the environment and performs the replanning process only on waypoints that are occupied by new obstacles. 
This helps to reduce the computation time and realize the proposed approach in real time. 
The feasibility of the proposed algorithm is evaluated using the Intel Aero Ready-to-Fly (RTF) quadcopter in simulation and in a real-world experiment.
\end{abstract}


\end{frontmatter}
\section{Introduction}
\label{section: introduction}
\input{introduction.tex}

\section{Modeling}
\input{chapter_2_modelling.tex}


\section{Two-step trajectory planning framework}
\label{section:Two-step trajectory (re)planning framework}
\input{chapter_3_trajectory_planning}

\section{Results}
\label{section: results}
\input{results.tex}

\section{Conclusions}
\label{section: Conclusion}
\input{conclusion.tex}

\bibliography{drone_IFAC2023.bib}
\end{document}

%% file: introduction.tex
Today, unmanned aerial vehicles (UAVs) are routinely used in many applications ranging from rapid delivery of goods, see \cite{singireddy2018technology}, and comprehensive sensing and surveillance of the environment, see, e.g., \cite{risbol2018lidar}, to various service delivery tasks, see, e.g., \cite{papachristos2019autonomous} and \cite{alwateer2020emerging}. The growth in computing power and battery capacity makes UAVs an interesting device in different fields. 
In particular, small and mechanically simple systems such as quadcopters are becoming the most popular flying robots. 
Compared to fixed-wing aircraft, such as airplanes, quadcopters are more maneuverable because they can hover in the air, making them more suitable for many applications.
For the autonomous operation of UAVs, the ability to quickly plan trajectories that move the quadcopter from an initial state to a final state is an important prerequisite.

Typically, the system state consists of position, velocity, orientation, and angular velocity. 
The planned trajectory must be dynamically feasible and meet additional conditions, such as avoiding collisions with obstacles. 
Quadcopter trajectory planning has been intensively researched on for years and can be divided into two main categories. 
In the first category, the motion primitive approach is the core concept, see \cite{lavalle2006planning}, where motions of robotic systems are computed to generate a motion library, also called motion primitives. 
Since each motion primitive is computed to be dynamically feasible, this approach is often used for online quadcopter replanning, see, e.g., \cite{andersson2018receding}, and \cite{liu2018search}. 
In \cite{pivtoraiko2013incremental}, an incremental replanning algorithm is proposed that uses offline motion primitives and reuses previous computations to produce a smooth and dynamically feasible trajectory. 
In addition, the use of offline motion primitives, also known as ``memory of motions'' is successfully applied in other systems such as the gantry crane system, see, e.g., \cite{vu2020fast} and \cite{vu2022fast}, and the collaborative robot system, see, e.g., \cite{vu2021fast}.
In \cite{mueller2015computationally}, a minimal jerk primitive is generated online with a given current state and a desired final state. 
Recently, the motion primitive approach was used to estimate probabilistic maneuvers for collision avoidance, see \cite{florence2020integrated}. 
In this approach, maneuver outputs are calculated based on unconstrained targets and collision avoidance. 
Although approaches based on motion primitives have been successfully employed in several applications, the construction of a motion primitive library is time-consuming. 
Moreover, adapting an already computed motion primitive library in case of a change in the corresponding environment is still a challenge. 

The second category includes a two-step approach consisting of a sampling-based trajectory planning algorithm and an optimization-based trajectory planning algorithm, see, e.g., \cite{gao2018online}.  
In the first step, sampling-based trajectory planning approaches, e.g., Rapidly exploring Random Tree (RRT*) in \cite{ramana2016motion} and \cite{vu2022sampling} or Fast Marching Method (FMT) in \cite{janson2015fast}, are used to generate a collision-free trajectory consisting of multiple waypoints. 
Since a quadcopter is a differentially flat system, where the position of the center of gravity and the yaw angle serve as flat outputs, see, e.g., \cite{fliess1995flatness} and \cite{faessler2017differential}, waypoints containing flat outputs are often considered in the first planning step. 
In the second step, optimization-based algorithms are used to generate a smooth and dynamically feasible trajectory through all computed waypoints, see, e.g., \cite{liu2016high}. 
By exploiting the flatness property of a quadcopter, all states and control inputs can be parameterized by the (sufficiently smooth) generated trajectory. 
In \cite{hehn2011quadrocopter}, the authors employ the RRT* to compute the waypoints of a quadcopter path. 
Then, a constrained quadratic program is used to solve the minimum-capture trajectory problem that fits all these waypoints into a polynomial. 
Other optimization-based approaches for the second step are presented in \cite{mehdi2015collision}, and \cite{gao2018online}, which optimize points of a B-splines trajectory. 
By modifying the path computed in the first step, the two-step approach can change the shape of the previously computed trajectory to avoid collisions with obstacles, see, e.g., \cite{tordesillas2021faster}.

In this work, the following scenario is considered. A quadcopter with a front-facing camera needs to move from an initial position to a target position in an indoor environment without any knowledge of the environment, e.g., obstacles. 
Inspired by the effectiveness of two-stage trajectory planning approaches, the focus of this work is to propose an online trajectory planning algorithm with the ability to re-plan in the event of the presence of new obstacles. 
Similar to approaches in the second category, RRT* is first used to compute a coarse collision-free path consisting of multiple waypoints. 
These waypoints consist of the position of the center of mass (CoM) of the quadcopter. 
This coarse path is then smoothed using the Line-of-Sight (LOS) algorithm, see \cite{naeem2012colregs}, which reduces the number of waypoints and increases the computational speed of the trajectory generation in the second step. The alignment of the quadcopter with the flight direction is done using separately computed yaw angle waypoints. 
Then, constrained quadratic programming is applied in the second step to compute a polynomial that passes through all computed waypoints. 
Once an obstacle is detected that collides with the current trajectory, a sub-algorithm is presented to quickly reconstruct a collision-free path and generate a new trajectory. 
Different from other works in the literature, only the nodes in the computed path that are encountered by the new object are recomputed.
This helps to speed up the computation of the subsequent trajectory generation process. 

The main contributions of this paper can be summarized as follows:
\begin{itemize}
    \item The LOS algorithm is applied in the first step to remove unnecessary waypoints. This modification helps to make the proposed algorithm real-time capable.
    \item An algorithm for detecting new obstacles using only a front-facing RGB-D camera is proposed, which is faster and less memory-consuming than classical approaches, see, e.g., \cite{dairi2018obstacle}. 
    \item The proposed online replanning algorithm is implemented in both simulations and real experiments using a companion computer and the Intel Aero RTF drone. 
\end{itemize}

The paper is organized as follows. Section 2 briefly introduces the differentially flat mathematical model of a quadcopter. In Section 3, the proposed algorithm for online trajectory replanning is presented. Section 4 presents the experimental setup, simulations, and experimental results. Finally, Section 5 concludes this paper and gives an outlook on future work.

%% file: chapter_2_modelling.tex
\begin{figure}
    \centering
    \includegraphics[height=.2\textheight]{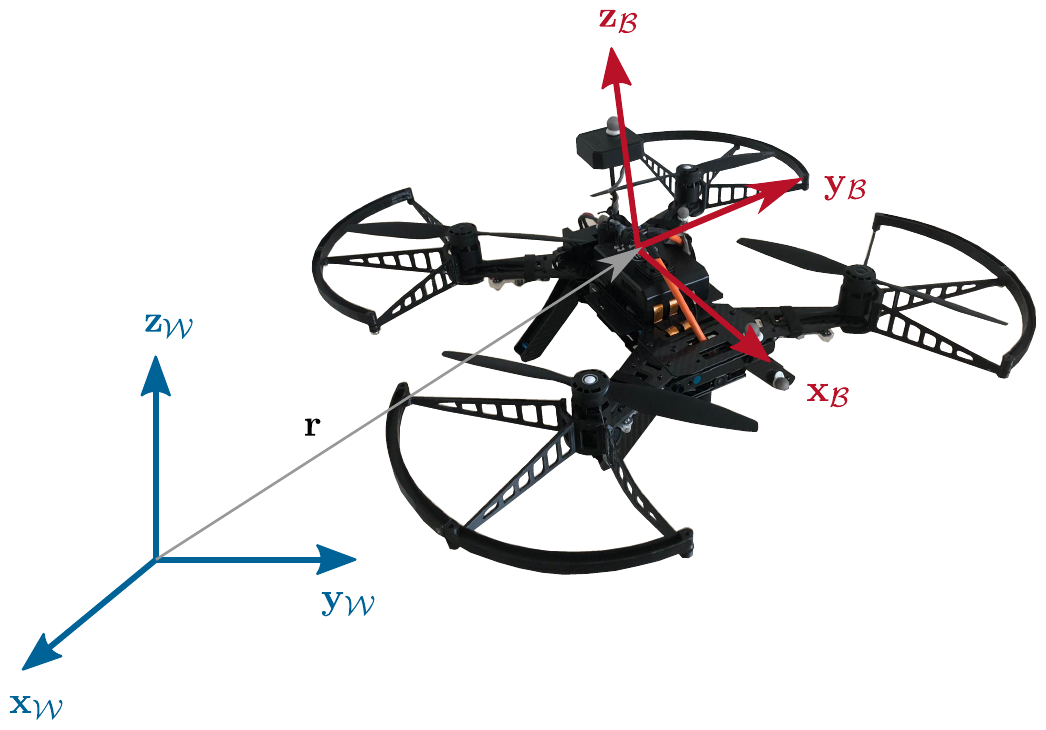}
    \caption{Coordinate systems of a quadcopter. The vector $\mathbf{r}$ denotes the position of the center of mass in the world coordinate.}
    \label{fig: UAVframes}
\end{figure}

This section briefly introduces the mathematical model as well as the differential flatness property of the quadcopter. 
The world coordinate system $\mathcal{W}$, consisting of three unit vectors $(x_\mathcal{W}\:, y_\mathcal{W},\: z_\mathcal{W})$, and the body coordinate system $\mathcal{B}$, consisting of three unit vectors $(x_\mathcal{B}\:, y_\mathcal{B}\: z_\mathcal{B})$, are shown in Fig. \ref{fig: UAVframes}. 
The $Z$-$X$-$Y$ Euler angle convention is used to express the rotation of the quadrotor in the world frame $\mathcal{W}$ in the form
\begin{equation}
    \mathbf{R}^\mathcal{W}_\mathcal{B} = \mathbf{R}_{z,\psi} \mathbf{R}_{x,\phi} \mathbf{R}_{y,\theta} \Comma
\end{equation}
where $\mathbf{R}_{i,\alpha}\:\mathrm{with}\: i \in \{x,y,z\}\:\mathrm{and} \: \alpha \in \{\psi,\:\phi,\:\theta\}$ denotes the rotation around the axis $i$ with the angle $\alpha$. 
The three angles $\psi$, $\phi$, and $\theta$ are also called yaw, roll and pitch, respectively. 
The angular velocity of the quadcopter's CoM in the world coordinate system $\bm{\omega}^{\mathcal{W}}_\mathcal{B} =[\omega_x, \omega_y, \omega_z]^\mathrm{T}$ is calculated from the skew-symmetric matrix operator $\mathbf{S}(\bm{\omega}^{\mathcal{W}}_\mathcal{B})$ as
\begin{equation}
\label{eq: eom 1}
    \mathbf{S}(\bm{\omega}^{\mathcal{W}}_\mathcal{B}) = \dfrac{\mathrm{d}}{\mathrm{d}t}({\mathbf{R}}_\mathcal{B}^\mathcal{W}) ({\mathbf{R}}_\mathcal{B}^\mathcal{W})^\mathrm{T} = \begin{bmatrix}
        0 & -\omega_z & \omega_y \\
        \omega_z & 0 &-\omega_x \\
        -\omega_y & \omega_x & 0
    \end{bmatrix}
    \: .
\end{equation}

\begin{figure}[htbp]
    \centering
    \def\svgwidth{\linewidth}
    \includegraphics[height=.2\textheight]{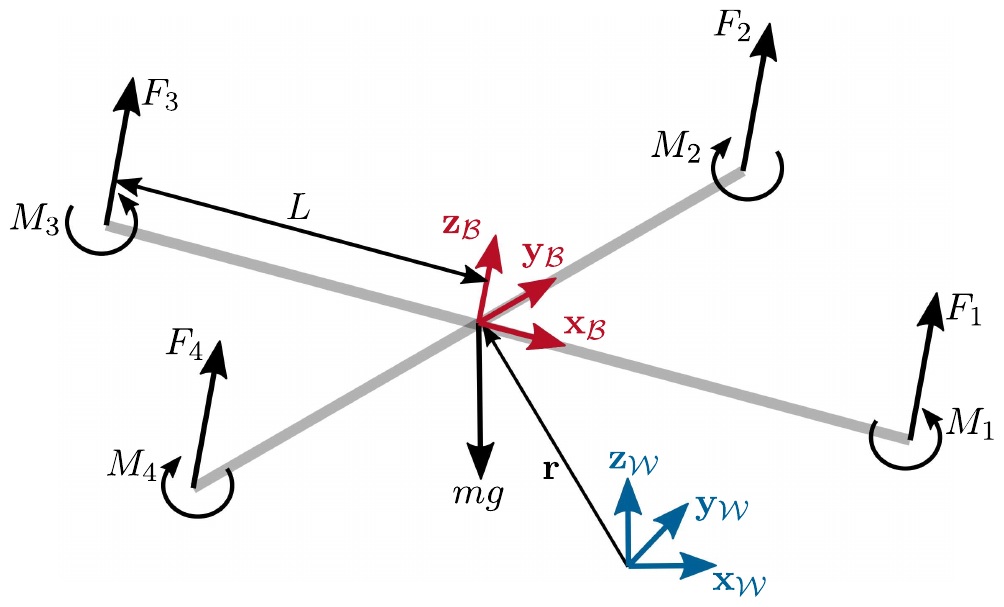}
    \caption{Free-body diagram of a quadcopter. $L$ is the length from the rotor to the CoM.}
    \label{fig:quadrotor_forces_moments}
\end{figure}
The free-body diagram describing all forces acting on a quadcopter is shown in Fig. \ref{fig:quadrotor_forces_moments}, where $F_i$ and $M_i$ with $i \in \{1,...,4\}$ are the force and moment exerted by the four rotors of the quadcopter. 
The control input $\mathbf{u}^\mathrm{T}= [u_1,\mathbf{u}^\mathrm{T}_2]$ of the system consists of the net thrust $u_1=\sum_{i=1}^4{F_i}$, and the moment vector 
\begin{equation}
    \vec{u}_2 = \begin{bmatrix}
         L(F_2 - F_4) \\ L(F_3 - F_1) \\ M_1 - M_2 + M_3 - M_4
    \end{bmatrix}.
    \label{eq:inputs}
\end{equation}

Using Newton's second law, equations for the linear acceleration of the CoM $\ddot{\mathbf{r}}$ and Euler's equations for the angular acceleration $\dot{\bm{\omega}}^{\mathcal{W}}_\mathcal{B}$, the equations of motion of the quadcopter can be written in the following form
\begin{equation}
    \begin{aligned}
       m\ddot{\mathbf{r}} &= [0,0,-mg]^\mathrm{T} + \mathbf{R}_\mathcal{B}^\mathcal{W} [0,0,u_1]^\mathrm{T} \\
        \dot{\bm{\omega}}^{\mathcal{W}}_\mathcal{B} &= \mathbf{I}^{-1}(-{\bm{\omega}}^{\mathcal{W}}_\mathcal{B} \times \mathbf{I}{\bm{\omega}}^{\mathcal{W}}_\mathcal{B}+\mathbf{u}_2^\mathrm{T})
        \Comma
        \label{eq: eom 2}
    \end{aligned}
\end{equation}
where $m$, $\mathbf{I}$, and $g$ are the mass, the moment of inertia matrix of the quadcopter, and the gravitational acceleration, respectively. 

Considering the state of the quadcopter system as $\mathbf{x}^\mathrm{T} = [\mathbf{r}^\mathrm{T},\dot{\mathbf{r}}^\mathrm{T},({\bm{\omega}}^{\mathcal{W}}_\mathcal{B})^\mathrm{T},(\dot{\bm{\omega}}^{\mathcal{W}}_\mathcal{B})^\mathrm{T}]$ and using (\ref{eq: eom 1}) and (\ref{eq: eom 2}), the equations of motion can be expressed in the state space form as
\begin{equation}
    \dot{\mathbf{x}} = \mathbf{f}(\mathbf{x},\mathbf{u}) \:\: \mathrm{with} \:\: \mathbf{u}^\mathrm{T} = [u_1,\mathbf{u}_2^\mathrm{T}]\: .
\end{equation}

A quadcopter is a differentially flat system with flat outputs chosen as its position $\mathbf{r}=[x,y,z]^\mathrm{T}$ and its yaw angle $\psi$, see, e.g., \cite{fliess1995flatness} and \cite{faessler2017differential}. The differential flatness property allows the state $\mathbf{x}$ and the control input $\mathbf{u} = [u_1,\mathbf{u}_2^{\mathrm{T}}]^\mathrm{T}$ to be parameterized by flat outputs and their time derivatives in the form
\begin{equation}
    ( \mathbf{x}, \mathbf{u} ) = \Phi ( \bm{\sigma}, \dot{\bm{\sigma}}, \ddot{\bm{\sigma}},{\bm{\sigma}}^{(3)},{\bm{\sigma}}^{(4)}) \Comma
    \label{eq: smoothmap}
\end{equation}
where $\bm{\sigma} = \begin{bmatrix} \mathbf{r}^\mathrm{T},\psi \end{bmatrix}^\mathrm{T}$. Due to the page restrictions, the derivation of (\ref{eq: smoothmap}) is omitted. 
For more information, the reader is referred to \cite{faessler2017differential}.

%% file: chapter_3_trajectory_planning.tex
Since the control inputs $\mathbf{u}$ can be computed from the flat outputs $\bm{\sigma}$ and their time derivatives by using (\ref{eq: smoothmap}), the proposed method plans a trajectory of the flat outputs $\bm{\sigma}$ from a given starting position $\mathbf{r}_{s}$ and a starting heading angle $\psi_s$ to a given target position $\mathbf{r}_{t}$ and a target heading angle $\psi_t$. 
In the following, the proposed trajectory planning framework is presented in detail.

\subsection{Overview of the proposed method}
\begin{figure}
    \centering
    \includegraphics[height=.4\textheight]{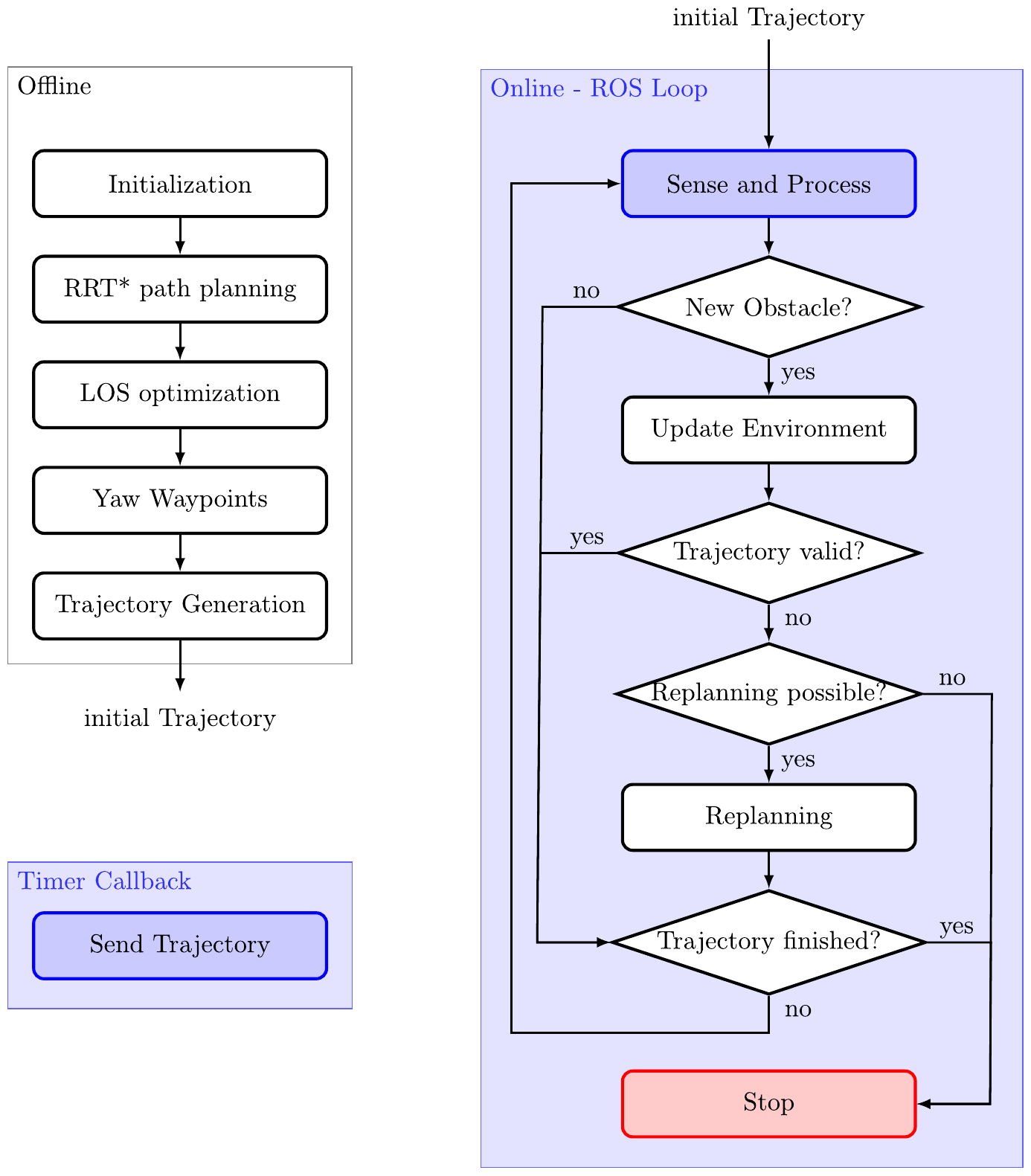}
    \caption{Overview of the proposed trajectory (re)planning framework.}
    \label{fig:TrajectoryPlanning}
\end{figure}

The proposed trajectory planning framework consists of an offline and online trajectory planning block and is depicted in Fig. \ref{fig:TrajectoryPlanning}. The offline block is executed at the beginning of a motion task.
The online block is executed while the quadcopter is moving to react to possible changes in the environment. 
In both blocks, the two-stage trajectory planning algorithm is included. 
However, in the online block, the two-stage trajectory planning algorithm is executed only for those subsets of the collision-free path computed in the offline block which is occupied by obstacles.    

In the first step of the offline block, the classical RRT$^*$ algorithm is used to compute a collision-free path of multiple waypoints from an initial position $\mathbf{r}_s$ to a target position $\mathbf{r}_t$. 
To reduce the redundancy of this path, the Line-of-Sight (LOS) algorithm is implemented to remove unnecessary waypoints. 
The Gilbert-Johnson-Keerthi (GJK) algorithm, see \cite{Gilbert:88}, is employed to calculate the Euclidean distances between the waypoints and the obstacles. 
The yaw waypoints are computed separately to ensure in-flight detection of possible objects along the planned path.

In the second step, the computed path is divided into a sequence of polynomial segments between waypoints, which are optimized into smooth trajectories using the constrained quadratic programming method proposed in \cite{Mellinger_thesis:2012} and \cite{Richter:16}. 
Then the online block on the right hand side of Fig. \ref{fig:TrajectoryPlanning} is executed. 
During the flight, the RGB-D camera is used to check for unknown obstacles in the environment. 
As soon as new obstacles are detected, a feasibility check of the precomputed trajectory is initiated, resulting in a recomputation of waypoints of the computed path and a new trajectory generation.

\subsection{First step: sampling-based approaches for computing a collision-free path}
The pseudo code of the RRT* to generate a tree $\mathcal{G}$ consisting of a set of nodes $\mathcal{V}$ and a set of edges $\mathcal{E}$ is presented in Alg. \ref{alg:RRTstar_alt}. 
The set $\mathcal{V}$ contains nodes $\{\mathbf{r}_1, ...,\mathbf{r}_N\}$ which are the CoM positions of the quadcopter in the collision-free space $\mathcal{R}_{free}$. 
$\mathcal{O}$ and $N$ denote the set of obstacles and the maximum size of the set of nodes $\mathcal{V}$, respectively. 
Additionally, the set of edges $\mathcal{E}$ contains the set of parent nodes of the corresponding nodes in the set $\mathcal{V}$. 
A parent node $\mathrm{Parent}(\mathbf{r})$ of the node $\mathbf{r}$ denotes the node that yields the least total distance to the target node $\mathbf{r}_t$.
When obtaining the target position $\mathbf{r}_t$ of the quadcopter, the RRT* algorithm starts to generate a node $\mathbf{r}_{rand}$ in $\mathcal{R}_{free}$ at random (line 4 in Alg. \ref{alg:RRTstar_alt}). 
Subsequently, the function $\mathrm{AddNode}$ adds this node to the tree $\mathcal{G}$ taking into account the set of obstacles $\mathcal{O}$ and two user-defined parameters, i.e., $\epsilon,\:\mathrm{and}\:\rho$. 
Note that $\mathbf{r}_{rand}$ is not directly added to the tree $\mathcal{G}$. 
Instead, the function $\mathrm{AddNode}$ selects the subset of the nodes in the proximity of the distance $\rho$ w.r.t. $\mathbf{r}_{rand}$. 
From this subset, the node $\mathbf{r}_{rand}^*$ which yields the smallest total distance to the target position $\mathbf{r}_t$ is chosen.
Finally, the tree $\mathcal{G}$ includes the node whose distance to $\mathbf{r}_{rand}^*$ is equal to $\epsilon$ and lies on the straight line between $\mathbf{r}_{rand}^*$ and $\mathbf{r}_{rand}$. 
The function $\mathrm{AddNode}$ is processed until the maximum size of the set $\mathcal{V}$ is reached. 
In lines 9-12 in Alg. \ref{alg:RRTstar_alt}, the shortest collision-free path from $\mathbf{r}_s$ to $\mathbf{r}_t$ is retrieved and stored in the set of waypoints $\mathcal{P}_\mathbf{r}$.
\begin{algorithm2e}
	\caption{RRT* algorithm} 
	\label{alg:RRTstar_alt}
	\KwIn{$\mathbf{r}_{s},\mathbf{r}_{t},\mathcal{O}, N, \epsilon, \rho$}
    \KwOut{$\mathcal{G} = \left( \mathcal{V},\mathcal{E} \right),\mathcal{P}_{\mathbf{r}}$}   
    
	$\mathcal{V} \leftarrow \{\mathbf{r}_{t}\}; \:\mathcal{E} \leftarrow \emptyset$
	
	$\mathcal{G} = \left( \mathcal{V},\mathcal{E} \right)$
	
	\While{$ \mathrm{size}(\mathcal{V}) \leq N$}{
	    $\mathbf{r}_{rand} \leftarrow \mathsf{Random}\left( \mathcal{R}_{free} \right)$
	    
	    $\mathcal{G} \leftarrow \mathsf{AddNode}\left(\mathbf{r}_{rand},\mathcal{G},  \mathcal{O}, \epsilon, \rho \right)$
	}
	$\mathcal{G} \leftarrow \mathsf{AddNode}\left(\mathbf{r}_{s}, \mathcal{G}, \mathcal{O}, \epsilon, \rho \right)$
	
	$\mathcal{P}_{\mathbf{r}} \leftarrow \{ \mathbf{r}_{s} \}$; $ \mathbf{r}_{temp} \leftarrow \mathbf{r}_{s}; $
 
        \While{$\mathsf{Parent}\left( \mathbf{r}_{temp} \right) \neq \mathbf{r}_{t}$}{
        $\mathbf{r}_{temp} \leftarrow \mathsf{Parent}\left( \mathbf{r}_{temp} \right)$
        
        $\mathcal{P}_{\mathbf{r}} \leftarrow \mathcal{P}_{\mathbf{r}} \cup \{ \mathbf{r}_{temp} \}$
	}
	$\mathcal{P}_{\mathbf{r}} \leftarrow \mathcal{P}_{\mathbf{r}} \cup \{ \mathbf{r}_{t} \}$
\end{algorithm2e}

Next, the Line-of-Sight (LOS) optimization, see Alg. \ref{alg:LOS}, is used to remove redundant nodes from the set of position waypoints $\mathcal{P}_{r}$. 
Note that $M$ is the size of the set of position waypoints $\mathcal{P}_\mathbf{r}$. 
Starting from the first node $\mathbf{r}_{in}=\mathcal{P}_\mathbf{r}(1)$, the LOS algorithm searches for the longest possible collision-free path from that node. 
If there is a collision-free path between the node $\mathbf{r}_{in}$ and $\mathcal{P}_\mathbf{r}(M-i)$, the function $\mathrm{Del\_wp}$ is used to delete redundant waypoints from these nodes, and the length of the set of waypoints $\mathcal{P}_\mathbf{r}$ is updated, see lines 3-7 in Alg. \ref{alg:LOS}. Otherwise, the counting index $i$ is incremented (line 8 in Alg. \ref{alg:LOS}). 
This process is repeated until the stopping criterion is met. 

\begin{algorithm2e}
	\caption{Line-of-Sight Optimization} 
	\label{alg:LOS}
	\KwIn{$\mathcal{P}_{\mathbf{r}},\mathcal{O}$}
    \KwOut{$\mathcal{P}_{\mathbf{r}}$}
	
	$\mathbf{r}_{in} \leftarrow \mathcal{P}_{\mathbf{r}}(1)$;\:\:$i=0$
	
	\While{$\mathbf{r}_{in} \not= \mathcal{P}_{\mathbf{r}}(M)$}{
	
	    \If{$ \mathsf{IsCollisionFree}(\mathbf{r}_{in},\mathcal{P}_{\mathbf{r}}(M-i),\mathcal{O})$}{
	        $\mathcal{P}_{\mathbf{r}}\leftarrow \mathsf{Del\_wp}(\mathbf{r}_{in},\mathcal{P}_{\mathbf{r}}(M-i))$

            $M=\mathrm{length}(\mathcal{P}_\mathbf{r})$
            
	        $\mathbf{r}_{in} \leftarrow \mathcal{P}_{\mathbf{r}}(M-i)$; \:$i=0$\:;
	    }
        $i = i + 1 $
	}
\end{algorithm2e}
Since the RRT* and LOS algorithms define only the position path $\mathcal{P}_\mathbf{r}$, the set of yaw waypoints $\mathcal{P}_\psi = \{\psi_1,...,\psi_M\}$ must be calculated accordingly. 
Given that the forward-facing camera is used in the experiments to detect obstacles, the yaw path is designed so that this camera points in the direction of flight. 
For this purpose, the yaw waypoints are calculated in the following form.
\begin{equation}
    \psi_i = \arctan \frac{r_{i+1,y} - r_{i,y}}{r_{i+1,x} - r_{i,x}}, i = {2,\dots,M-1} \Comma
    \label{eq:yaw_angle_def}
\end{equation}
where $r_{i,x}$ and $r_{i,y}$ are the first and second components of the vector $\mathbf{r}_i$. 
Note that the start $\psi_1$ and the target yaw angle $\psi_M$ are predefined by the user. 
Here, the collision-free path of the flat output $\mathcal{P}_{\bm{\sigma}}$ is found by the set of position nodes $\mathcal{P}_\mathbf{r}$ and the set of yaw nodes $\mathcal{P}_\psi$. 
\subsection{Second step: Constrained quadratic programming for trajectory generation}
Similar to \cite{Mellinger_thesis:2012}, the trajectory generation is split into four independent optimization problems for each flat output parameterized by the time $t$. 
A common notation $P(t)$ is used for each flat output in $\mathcal{P}_{\bm{\sigma}}$. 
The flat output trajectory $P(t)$ of $M$ waypoints is given by a piecewise polynomial function of $M-1$ segments in the form 
\begin{equation}
    \begin{aligned}
    &P(t) = \bigg\{P_j(t)=\sum_{i=0}^n p_{i,j}t^i \bigg\}\\
    & \:j\in \{1,\dots,M-1\},\:\:t_{j}\leq t < t_{j+1} \Comma
    \end{aligned}    
\end{equation}
where $[t_{j},\:t_{j+1})$ is the time segment, $n$ is the order of the trajectory polynomials, and $p_{i,j}$ are the coefficients of the polynomial $P_j(t)$. 
The cost function for the $j^{th}$ segment reads as
\begin{equation}
    \begin{aligned}
    J_j &= \int \limits_{t_j} \limits^{t_j+T_j} \sum_{i=1}^n w_i \bigg[\dfrac{\mathrm{d}^iP(t)}{\mathrm{d}t^i}\bigg]^2 \mathrm{d}t
    = \mathbf{p}_j^T\mathbf{Q}_j\mathbf{p}_j,
    \end{aligned}
\label{eq:costfunctional_segment}
\end{equation}
where $\mathbf{p}_j$ is the vector of the $n+1$ coefficients of the polynomial $P_j(t)$, $w_i > 0$ is the user-defined weight of the $i^{th}$ derivative, and $T_j = t_{j+1}-t_j$ is the segment time. 
Note that $\mathbf{Q}_j$ is a positive definite matrix. 
The segment times $T_j$ can be chosen heuristically by a desired average velocity within the segment. 
To guarantee the (sufficient) smoothness of the flat output trajectory, constraints on the endpoints of each segment $j$ are considered via a mapping matrix $\mathbf{A}_j$ between the polynomial coefficients $\mathbf{p}_j$ and the time derivatives $\mathbf{d}_j$ at the endpoints in the form
\begin{equation}
    \mathbf{A}_j\mathbf{p}_j = \mathbf{d}_j\:.
    \label{eq:constraint_equ}
\end{equation}
All cost functions $J_j$ in (\ref{eq:costfunctional_segment}) and constraints in (\ref{eq:constraint_equ}) of the $M$ segments are combined, which leads to the following constrained optimization problem

\begin{equation}
\begin{aligned}
&\min_{\mathbf{p}_1,\ldots,\mathbf{p}_{M-1}} \quad \sum_{j=1}^{M - 1}J_j\\
&\textrm{s.t.} \:\:   \begin{bmatrix}
    \mathbf{A}_1 & 0 & \dots & 0 \\
    0 & \mathbf{A}_2 & \dots & 0 \\
    \vdots & \vdots & \ddots & \vdots \\
    0 & 0 & \dots & \mathbf{A}_{M-1}
  \end{bmatrix}\begin{bmatrix}
        \mathbf{p}_1 \\ \mathbf{p}_2 \\ \vdots \\ \mathbf{p}_{M-2}\\ \mathbf{p}_{M-1}
    \end{bmatrix}
    = \begin{bmatrix}
        \mathbf{d}_1 \\ \mathbf{0} \\ \vdots \\\mathbf{0}\\ \mathbf{d}_{M-1}
    \end{bmatrix} \:.
\end{aligned}
\end{equation}
\subsection{Trajectory (re)planning}
During the flight, the $\mathrm{Online-ROS \: Loop}$, see, Fig. \ref{fig:TrajectoryPlanning}, is activated to take into account changes in the environment using the block $\mathrm{Sense \: and \: Process}$ and generate the new trajectory using the block $\mathrm{Replanning}$.
Obstacles in the set $\mathcal{O}$ are considered to be convex, i.e. cuboids. 
The point clouds of the front camera are filtered in the block $\mathrm{Sense}$ $\mathrm{and}$ $\mathrm{Process}$.
The pseudocode of the eight-corner obstacle detection approach is shown in Alg. \ref{alg:eight-corner}. 

First, the captured point cloud data is clustered and converted into 8-corner boxes using the function $\mathrm{convertPc2Box}$ (line 2 in Alg. \ref{alg:eight-corner}). 
If the current set of obstacles is empty, all clustered 8-corner boxes (cuboids) are added as new obstacles (lines $3-6$ in Alg. \ref{alg:eight-corner}). 
Otherwise, the function $\mathrm{getDistanceMatrix}(\mathcal{C},\mathbf{O})$ is employed which computes a matrix containing the minimum distance between each detected cuboid $\mathbf{C} \in \mathcal{C}$ w.r.t. each known obstacle $\mathbf{O} \in \mathcal{O}$ (line $7$ in Alg. \ref{alg:eight-corner}). Here, if the smallest distance $d$ between $\mathbf{C}$ and the closest known obstacles in $\mathcal{O}$ is greater than the threshold value $\delta$, the cuboid $C$ is considered as a new obstacle (lines $9-11$ of Alg. \ref{alg:eight-corner}). 
When the clustered cuboid $C \in \mathcal{C}$ intersects with one or more known obstacles in $\mathcal{O}$, they are both merged (lines $12-15$ in Alg. \ref{alg:eight-corner}). 
To further remove redundancy due to the measurement noise, the distance of each corner of the newly clustered cuboid $\mathbf{C} \in \mathcal{C}$ to the closest known obstacle $\mathbf{O} \in \mathcal{O}$ is computed and analyzed (lines $16$-$20$, Alg. \ref{alg:eight-corner}). If the minimum distance $d$ of $\mathbf{C}$ w.r.t. a known obstacle $\mathbf{O} \in \mathcal{O}$ is smaller than $\delta$, and one corner point of $\mathbf{C}$ is further away than the threshold, both cuboids are merged. This helps to merge falsely detected clusters which are surfaces of a real obstacle. 
The falsely detected clusters are, among others, caused by camera noise, lighting conditions, and transformation errors due to measurement accuracy.

\begin{algorithm2e}
	\caption{8-corner obstacle detection.} 
	\label{alg:eight-corner}
	\KwIn{$\mathcal{O},clusters, \delta$}
    \KwOut{$\mathcal{O}_{new}, \mathcal{O}$}
	
	$\mathcal{O}_{new} \leftarrow \{ \}$
	
	$\mathcal{C} \leftarrow \mathsf{convertPc2Box}(clusters)$
	
	\If{$\left( \mathsf{size}(\mathcal{O}) == 0 \right) \And \left( \mathsf{size}(\mathcal{C}) > 0 \right)$}{
	    $\mathcal{O}_{new} \leftarrow \mathcal{C}$
	}
	
	$\mathbf{D}, \mathbf{I} \leftarrow \mathsf{min}\big( \mathsf{getDistanceMatrix}(\mathcal{C},\mathcal{O}) \big)$
	
	\For{$ \mathbf{C} \in \mathcal{C}$}{
	    
	    $d \leftarrow \mathbf{D}(i); idx \leftarrow \mathbf{I}(i);$
	    
	    \If{$d > \delta$}{
	        $\mathcal{O}_{new} \leftarrow \mathcal{O}_{new} \cup \mathbf{C}$
	    }
	    
	    \If{$d == 0$}{
	        $\mathcal{O}_{new} \leftarrow \mathcal{O}_{new} \cup \mathsf{mergeBoxes}(\mathbf{C}, \mathcal{O}(idx))$
	        
	        $\mathcal{O} \leftarrow \mathcal{O} \setminus \mathcal{O}(idx)$
	    }
	    
	    $\mathbf{d}_{corner} \leftarrow \mathsf{getCornerDistances}(\mathbf{C}, \mathcal{O}(idx))$
	    
	    \If{$\left(\mathsf{max}\left(\mathbf{d}_{corner}\right) > \delta \right)\And \left( d < \delta \right) $}{
	        $\mathcal{O}_{new} \leftarrow \mathcal{O}_{new} \cup \mathsf{mergeBoxes}(\mathbf{C}, \mathcal{O}(idx))$
	        
	        $\mathcal{O} \leftarrow \mathcal{O} \setminus \mathcal{O}(idx)$
	    }
	}
	
\end{algorithm2e}

Once new obstacles are detected, the block $\mathrm{Replanning}$ activates the RRT* algorithm for those parts of the computed tree $\mathcal{G}$ which are occupied by new obstacles. 
Once a new collision-free path is obtained from the LOS algorithm, two additional position waypoints are inserted into the first two segments of the path. In detail, if there are two or more path segments from the current position of the quadcopter to the target, the first two segments are bisected via additional position waypoints, which guide the polynomial-shaped trajectory more strictly past the previously recognized object.
This prevents the final position trajectory from overshooting at higher speeds. 
Moreover, a faster alignment of the heading of the quadcopter is achieved, which ensures the recognition of possible further objects in the flight path. 
Finally, the yaw waypoints are defined and trajectory optimization including the current initial state of the quadcopter is executed, yielding a smooth collision-free trajectory to the target.

%% file: results.tex
\begin{figure}
    \centering
    \includegraphics[height=.33\textheight]{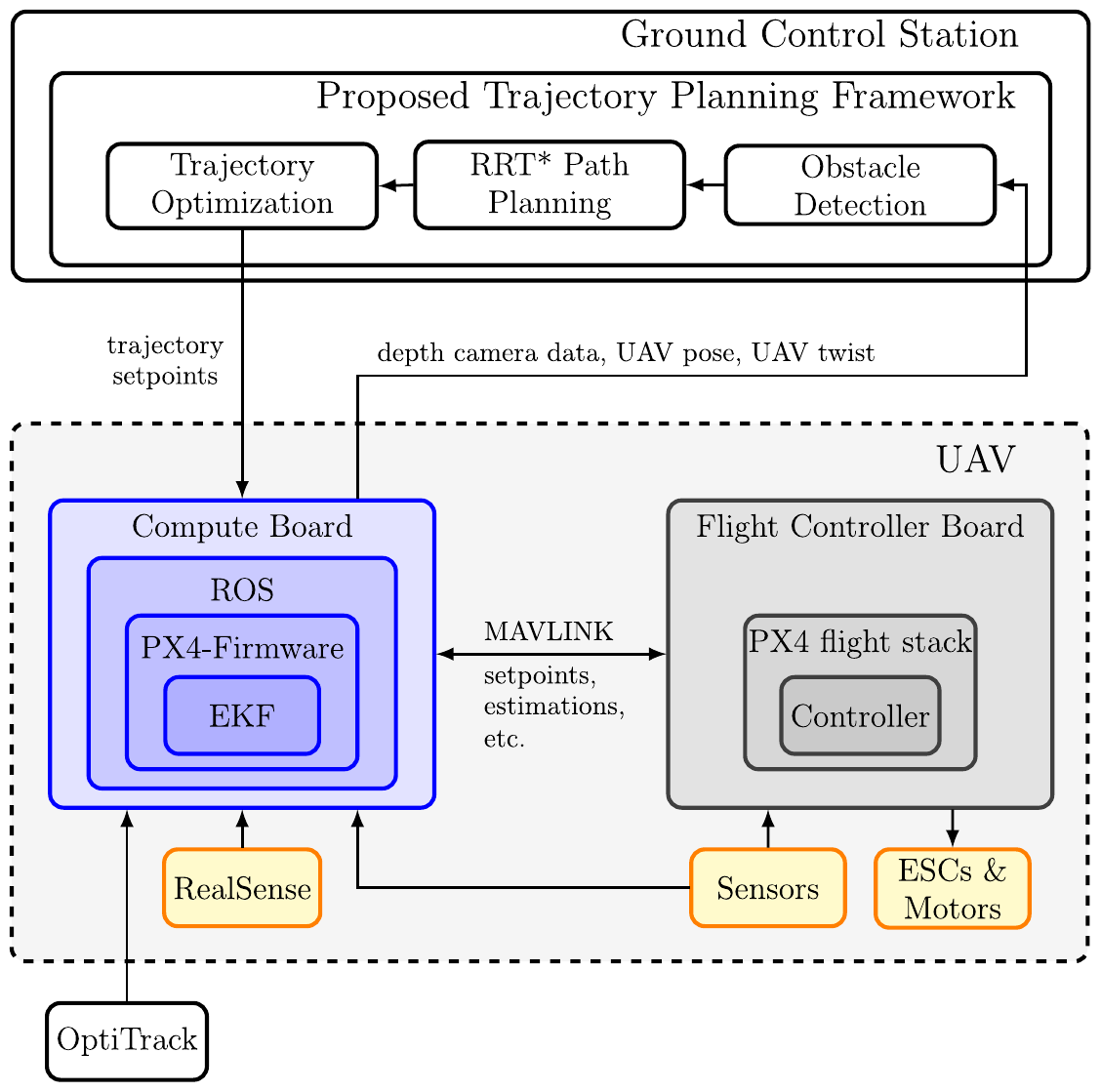}
    \caption{Overview of the simulation and experiment setup.}
    \label{fig:SystemArchitecture}
\end{figure}

To verify the feasibility of the proposed trajectory replanning framework in experiments, the Intel Aero RTF drone equipped with a front-facing RealSense R200 RGB-D camera is used. 
This quadcopter consists of two main components, the compute board running the Robot Operating System (ROS) middleware and the flight controller board running the PX4 flight stack, see Fig. \ref{fig:SystemArchitecture}. 
The RealSense R200 camera is used to acquire point cloud data for obstacle detection. 
The proposed online trajectory (re)planning is processed on a laptop, also named the Ground Control Station, with $1.8$ GHz Intel Core i7 and 16 GB RAM. The calculated trajectory is sent to the quadcopter via Wi-Fi at a rate of $6$ Hz. 

For the simulations, the quadcopter model is created in the open-source simulator Gazebo. 
This simulator provides a link to ROS and enables software-in-the-loop (SITL) simulation of the PX4 flight stack, which is also the firmware of the Intel Aero RTF drone. 
In addition, ROS provides a simulator for the RealSense R200 camera. 
\subsection{Simulation results}\label{sec:Simulation}
\begin{figure}[htbp]
	\centering
	\includegraphics[height=0.30\textheight]{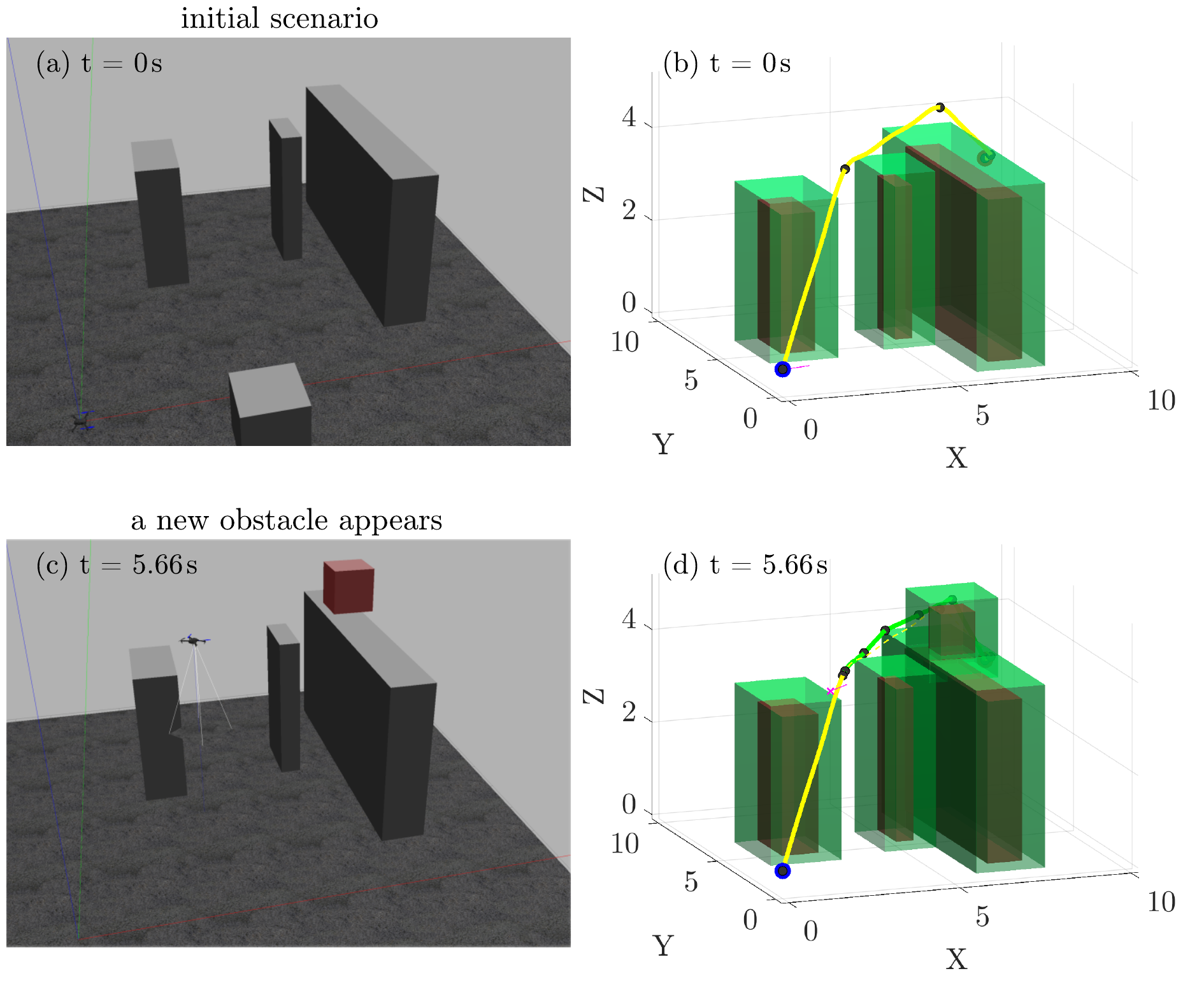}
    \caption{
     Snapshots of the collision-free path in the simulation. (a) The captured environment in Gazebo at $t = $\SI{0}{\second} with 3 obstacles. (b) Collision-free path computed from the $\mathrm{offline}$ block in Fig. \ref{fig:TrajectoryPlanning}. (c) The detected environment in Gazebo at $t = \SI{5.66}{\second}$ when a new obstacle appears (in red color). (d) Collision-free path computed from the $\mathrm{online}$ block in Fig. \ref{fig:TrajectoryPlanning} considering the new obstacle. 
     }
    \label{fig:SIM_Collage}
\end{figure}

\begin{figure}[htbp]
	\centering
	\includegraphics[height=0.56\textheight]{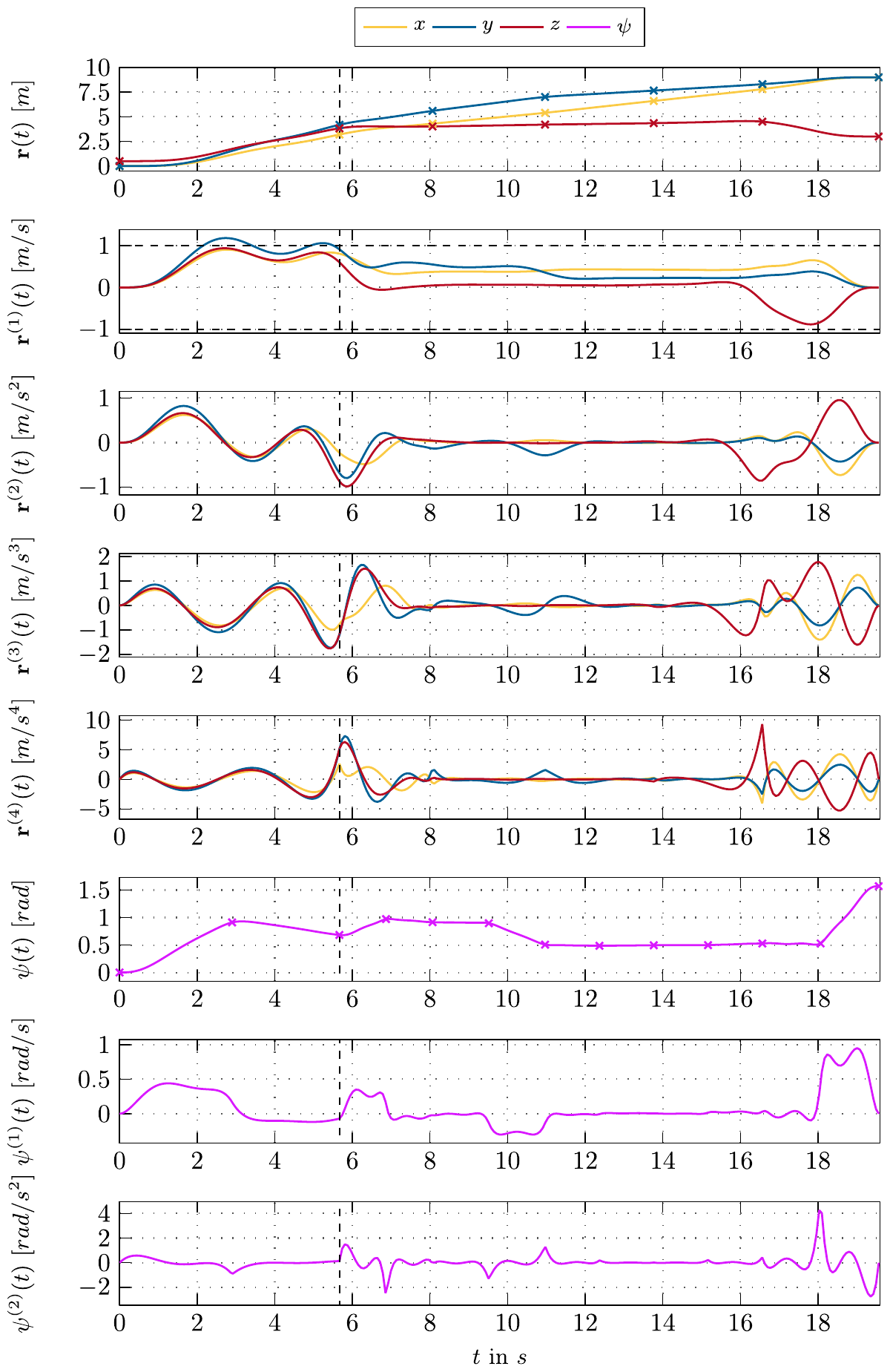}
    \caption{Time evolution of the generated trajectory in the simulation. The black vertical dashed line marks the time when the trajectory replanning is activated at \SI{5.66}{s}. The asterisks indicate the waypoints of the calculated trajectory and the upper index $(\mathrm{i})$ refers to the $i^\mathrm{th}$ time derivative of the corresponding quantity.}
    \label{fig:SIM_TrajectoryPlot}
\end{figure}

\begin{figure}[htbp]
	\centering
	\includegraphics[height=0.22\textheight]{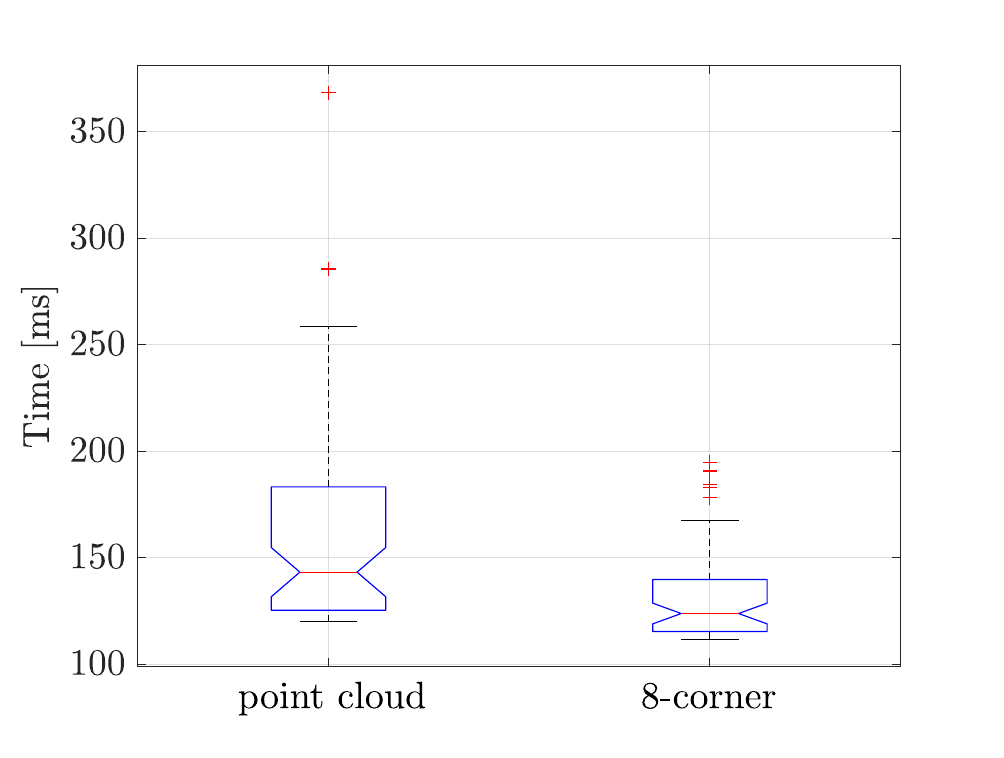}
    \caption{Monte Carlo simulations of the run times of obstacle detection approaches.}
    \label{fig:OBS_runtime}
\end{figure}

Fig. \ref{fig:SIM_Collage} illustrates snapshots of the simulation environment and the corresponding collision-free paths computed with the proposed algorithm. 
The quadcopter takes off from the lower left corner of the flight space, see Fig. \ref{fig:SIM_Collage}(a).  
The real sizes of the obstacles are shown in Fig. \ref{fig:SIM_Collage}(a) and (c), while the corresponding inflated obstacles are depicted in Fig.~ \ref{fig:SIM_Collage}(b) and (d) are considered in the proposed algorithm for safety reasons. 
Initially, $t=\SI{0}{\second}$, the quadcopter scans the environment and computes a collision-free path shown in Fig. \ref{fig:SIM_Collage}(b) in yellow. 
As soon as a new obstacle appears at $t= \SI{5.66}{\second}$ (shown in Fig. \ref{fig:SIM_Collage}(c) in red), the proposed trajectory replanning algorithm quickly computes the new collision-free trajectory, as shown in Fig. \ref{fig:PRES_Collage}(d) in green color.
The time evolution of the calculated trajectory in the simulation is shown in Fig. \ref{fig:SIM_TrajectoryPlot}. The waypoints are marked by asterisks. 
Overall, the trajectory generation results in a smooth trajectory that passes all waypoints. 
Moreover, smooth transitions are achieved at $t=\SI{5.66}{\second}$ for all flat outputs, cf. Fig. \ref{fig:SIM_TrajectoryPlot}.
Since obstacle detection plays an important role in the proposed online trajectory replanning algorithm, the comparison between the proposed 8-corner method and the classical point cloud method is investigated using Monte Carlo simulations. 
Different from the proposed 8-corner method, the classical point cloud method uses the $k$-nearest neighbor ($k$-NN) search on the clustered point cloud data, see \cite{pinkham2020quicknn}, to find the distances of the newly discovered clusters to the already known clusters. 
Then, this method decides whether a newly discovered cluster is an additional obstacle or not based on the corresponding distance. 
Instead of storing only 8 corners of a convex obstacle, this classical point cloud method processes all point cloud data. 
In Fig. \ref{fig:OBS_runtime}, the run times of the two approaches for the scenario in Fig. \ref{fig:SIM_Collage} over $81$ frames were calculated statistically. 
The results show that the 8-corner approach is faster than the classical point cloud approach. 
Moreover, the standard deviation of the computing time ($\approx\SI{83}{ms}$) of the proposed approach is much smaller compared to the standard deviation of the computing time ($\approx\SI{248}{ms}$) of the point cloud method. 
However, the 8-corner approach suffers from the possible false fusion of multiple objects in the presence of poor position estimates.  

\subsection{Experimental results}\label{sec:Experiment}
\begin{figure}[htbp]
	\centering
	\includegraphics[height=0.35\textheight]{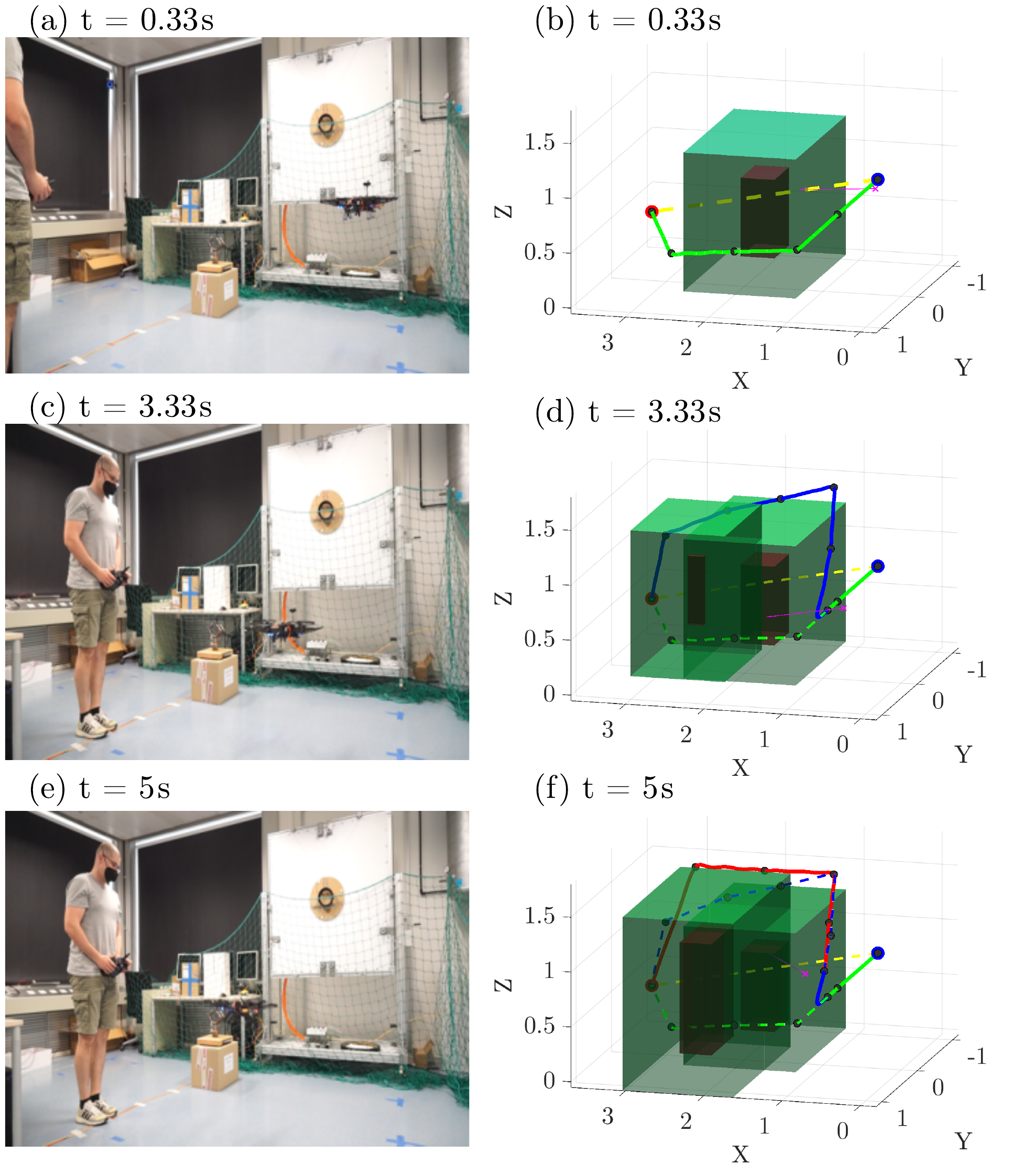}
    \caption{Snapshots of the experiment. The computed trajectories and detected obstacles with time stamps are illustrated on the right-hand side. 
    }
    \label{fig:PRES_Collage}
\end{figure}

\begin{figure}[htbp]
	\centering
	\includegraphics[height=.56\textheight]{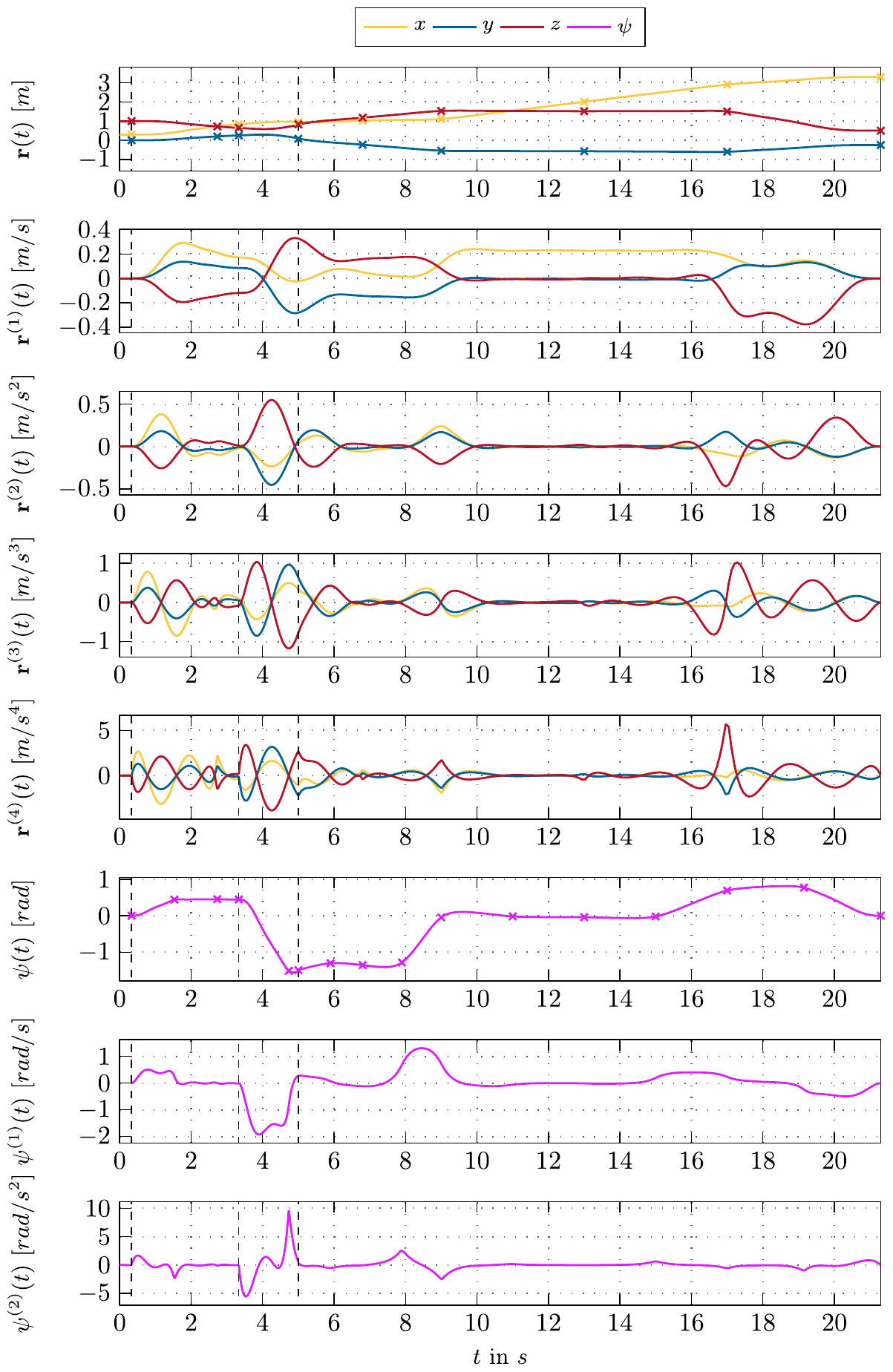}
    \caption{Time evolution of generated trajectory in the experiment. The black vertical dashed lines mark when the trajectory planning is activated. Asterisks indicate waypoints of the computed trajectory and the upper index $(\mathrm{i})$ refers to the $i^\mathrm{th}$ time derivative of the corresponding quantity.}
    \label{fig:PRES_TrajectoryPlot}
\end{figure}

In Fig. \ref{fig:PRES_Collage}(a), the flight environment of size $\SI{3.5}{\meter}$$\times$$\SI{2.5}{\meter}$$\times$$\SI{2}{\meter}$ contains a fixed obstacle on the ground. 
Since the obstacle is located between the start and end positions, a collision-free trajectory is calculated, shown as a green line in Fig. \ref{fig:PRES_Collage}(b) at $t=\SI{0.33}{\second}$. 
Then the quadcopter follows this trajectory until the human operator steps into the path. 
This activates the replanning algorithm that leads to the new collision-free path shown in green in Fig. \ref{fig:PRES_Collage}(d) at $t = \SI{3.33}{\second}$. 
When the RealSense camera captures the operator, the merging mechanism of the proposed algorithm is processed step-wise, resulting in an additional replanning step. 
Although the camera's viewpoint cannot capture the entire scene, the replanning capability helps to safely navigate the quadcopter to the target position. 
The time evolution of the computed trajectory in the experiment is shown in Fig. \ref{fig:PRES_TrajectoryPlot}. 
All derivatives of each flat output are continuous at the replanning time.
In the time between $\SI{3.33}{s}$ and $\SI{5}{s}$, a very sharp yaw maneuver takes place, the drone has to rotate around its own axis in a very short time to be reoriented in the flight direction. This results in a high yaw rate and strong spikes in angular acceleration. 
However, the quadcopter controller limits the yaw rate by default to increase safety. A video of the presented experiment and other scenarios is available at \url{https://www.acin.tuwien.ac.at/en/9d0b/}.

%% file: conclusion.tex
In this paper, a two-step online trajectory planning algorithm is proposed that can autonomously navigate the quadcopter from a start position to an end position in the presence of unknown obstacles using only a forward-facing RGB-D camera. 
In the first step, a collision-free path is computed by using the optimal Rapidly exploring Random Tree (RRT*) and the Line-of-Sight (LOS) optimization. 
Then, a trajectory is generated using constrained quadratic programming. 
Simulations and experiments verify the effectiveness of the proposed algorithm. 
The results show that the proposed algorithm is able to compute collision-free trajectories with sufficiently smooth system inputs. 
When unknown obstacles are detected in the flight path, a new collision-free trajectory is planned. 
The LOS algorithm plays an important role in reducing the number of waypoints of the path computed with the RRT* algorithm, which curbs the computational effort for the subsequent trajectory optimization. 
Moreover, the proposed 8-corner approach for obstacle detection is faster compared to the classical point cloud method. 

To improve the performance and quality of the proposed approach, several aspects can be considered in future work. First, the inclusion of dynamic constraints on the yaw angle in the estimation of segment times could ensure sufficient time for the drone to align with the flight path. In addition, the bending angle between successive edges can be considered in the cost of the RRT* algorithm to obtain a better linear path from the start point to the end point.

%% file: root.bbl
\begin{thebibliography}{29}
\providecommand{\natexlab}[1]{#1}
\providecommand{\url}[1]{\texttt{#1}}
\providecommand{\urlprefix}{URL }
\expandafter\ifx\csname urlstyle\endcsname\relax
  \providecommand{\doi}[1]{doi:\discretionary{}{}{}#1}\else
  \providecommand{\doi}{doi:\discretionary{}{}{}\begingroup
  \urlstyle{rm}\Url}\fi

\bibitem[{Alwateer and Loke(2020)}]{alwateer2020emerging}
Alwateer, M. and Loke, S.W. (2020).
\newblock Emerging drone services: Challenges and societal issues.
\newblock \emph{IEEE Technology and Society Magazine}, 39(3), 47--51.

\bibitem[{Andersson et~al.(2018)Andersson, Ljungqvist, Tiger, Axehill, and
  Heintz}]{andersson2018receding}
Andersson, O., Ljungqvist, O., Tiger, M., Axehill, D., and Heintz, F. (2018).
\newblock Receding-horizon lattice-based motion planning with dynamic obstacle
  avoidance.
\newblock In \emph{Proceedings of the Conference on Decision and Control},
  4467--4474.

\bibitem[{Dairi et~al.(2018)Dairi, Harrou, Sun, and
  Senouci}]{dairi2018obstacle}
Dairi, A., Harrou, F., Sun, Y., and Senouci, M. (2018).
\newblock Obstacle detection for intelligent transportation systems using deep
  stacked autoencoder and $ k $-nearest neighbor scheme.
\newblock \emph{IEEE Sensors Journal}, 18(12), 5122--5132.

\bibitem[{Faessler et~al.(2017)Faessler, Franchi, and
  Scaramuzza}]{faessler2017differential}
Faessler, M., Franchi, A., and Scaramuzza, D. (2017).
\newblock Differential flatness of quadrotor dynamics subject to rotor drag for
  accurate tracking of high-speed trajectories.
\newblock \emph{IEEE Robotics and Automation Letters}, 3(2), 620--626.

\bibitem[{Fliess et~al.(1995)Fliess, L{\'e}vine, Martin, and
  Rouchon}]{fliess1995flatness}
Fliess, M., L{\'e}vine, J., Martin, P., and Rouchon, P. (1995).
\newblock Flatness and defect of non-linear systems: introductory theory and
  examples.
\newblock \emph{International Journal of Control}, 61(6), 1327--1361.

\bibitem[{Florence et~al.(2020)Florence, Carter, and
  Tedrake}]{florence2020integrated}
Florence, P., Carter, J., and Tedrake, R. (2020).
\newblock Integrated perception and control at high speed: Evaluating collision
  avoidance maneuvers without maps.
\newblock In \emph{Algorithmic Foundations of Robotics XII}, 304--319.
  Springer, Cham, Switzerland.

\bibitem[{Gao et~al.(2018)Gao, Wu, Lin, and Shen}]{gao2018online}
Gao, F., Wu, W., Lin, Y., and Shen, S. (2018).
\newblock Online safe trajectory generation for quadrotors using fast marching
  method and {Bernstein} basis polynomial.
\newblock In \emph{Proceedings of the International Conference on Robotics and
  Automation}, 344--351.

\bibitem[{Gilbert et~al.(1988)Gilbert, Johnson, and Keerthi}]{Gilbert:88}
Gilbert, E., Johnson, D., and Keerthi, S. (1988).
\newblock A fast procedure for computing the distance between complex objects
  in three-dimensional space.
\newblock \emph{IEEE Journal on Robotics and Automation}, 4, 193--203.

\bibitem[{Hehn and D'Andrea(2011)}]{hehn2011quadrocopter}
Hehn, M. and D'Andrea, R. (2011).
\newblock Quadrocopter trajectory generation and control.
\newblock \emph{IFAC Proceedings Volumes}, 44(1), 1485--1491.

\bibitem[{Janson et~al.(2015)Janson, Schmerling, Clark, and
  Pavone}]{janson2015fast}
Janson, L., Schmerling, E., Clark, A., and Pavone, M. (2015).
\newblock Fast marching tree: A fast marching sampling-based method for optimal
  motion planning in many dimensions.
\newblock \emph{The International journal of robotics research}, 34(7),
  883--921.

\bibitem[{LaValle(2006)}]{lavalle2006planning}
LaValle, S.M. (2006).
\newblock \emph{Planning algorithms}.
\newblock Cambridge university press, Cambridge, United Kingdom.

\bibitem[{Liu et~al.(2018)Liu, Mohta, Atanasov, and Kumar}]{liu2018search}
Liu, S., Mohta, K., Atanasov, N., and Kumar, V. (2018).
\newblock {Search-based motion planning for aggressive flight in SE (3)}.
\newblock \emph{IEEE Robotics and Automation Letters}, 3(3), 2439--2446.

\bibitem[{Liu et~al.(2016)Liu, Watterson, Tang, and Kumar}]{liu2016high}
Liu, S., Watterson, M., Tang, S., and Kumar, V. (2016).
\newblock High speed navigation for quadrotors with limited onboard sensing.
\newblock In \emph{Proceedings of the International Conference on Robotics and
  Automation}, 1484--1491.

\bibitem[{Mehdi et~al.(2015)Mehdi, Choe, Cichella, and
  Hovakimyan}]{mehdi2015collision}
Mehdi, S.B., Choe, R., Cichella, V., and Hovakimyan, N. (2015).
\newblock Collision avoidance through path replanning using b{\'e}zier curves.
\newblock In \emph{Proceedings of the AIAA Guidance, navigation, and control
  conference}, 0598.

\bibitem[{Mellinger(2021)}]{Mellinger_thesis:2012}
Mellinger, D.W. (2021).
\newblock \emph{Trajectory {Generation} and {Control} for {Quadrotors}}.
\newblock Ph.D. thesis, University of Pennsylvania, , [Online: Accessed on 11
  November 2022].

\bibitem[{Mueller et~al.(2015)Mueller, Hehn, and
  D'Andrea}]{mueller2015computationally}
Mueller, M.W., Hehn, M., and D'Andrea, R. (2015).
\newblock A computationally efficient motion primitive for quadrocopter
  trajectory generation.
\newblock \emph{IEEE Transactions on Robotics}, 31(6), 1294--1310.

\bibitem[{Naeem et~al.(2012)Naeem, Irwin, and Yang}]{naeem2012colregs}
Naeem, W., Irwin, G.W., and Yang, A. (2012).
\newblock Colregs-based collision avoidance strategies for unmanned surface
  vehicles.
\newblock \emph{Mechatronics}, 22(6), 669--678.

\bibitem[{Papachristos et~al.(2019)Papachristos, Kamel, Popovi{\'c}, Khattak,
  Bircher, Oleynikova, Dang, Mascarich, Alexis, and
  Siegwart}]{papachristos2019autonomous}
Papachristos, C., Kamel, M., Popovi{\'c}, M., Khattak, S., Bircher, A.,
  Oleynikova, H., Dang, T., Mascarich, F., Alexis, K., and Siegwart, R. (2019).
\newblock Autonomous exploration and inspection path planning for aerial robots
  using the robot operating system.
\newblock In \emph{Robot Operating System (ROS)}, 67--111. Springer: Cham,
  switzerland.

\bibitem[{Pinkham et~al.(2020)Pinkham, Zeng, and Zhang}]{pinkham2020quicknn}
Pinkham, R., Zeng, S., and Zhang, Z. (2020).
\newblock Quicknn: Memory and performance optimization of kd tree based nearest
  neighbor search for 3d point clouds.
\newblock In \emph{Proceedings of the International Symposium on High
  Performance Computer Architecture}, 180--192.

\bibitem[{Pivtoraiko et~al.(2013)Pivtoraiko, Mellinger, and
  Kumar}]{pivtoraiko2013incremental}
Pivtoraiko, M., Mellinger, D., and Kumar, V. (2013).
\newblock Incremental micro-uav motion replanning for exploring unknown
  environments.
\newblock In \emph{Proceedings of the International Conference on Robotics and
  Automation}, 2452--2458.

\bibitem[{Ramana et~al.(2016)Ramana, Varma, and Kothari}]{ramana2016motion}
Ramana, M., Varma, S.A., and Kothari, M. (2016).
\newblock Motion planning for a fixed-wing uav in urban environments.
\newblock \emph{IFAC-PapersOnLine}, 49(1), 419--424.

\bibitem[{Richter et~al.(2016)Richter, Bry, and Roy}]{Richter:16}
Richter, C., Bry, A., and Roy, N. (2016).
\newblock Polynomial trajectory planning for aggressive quadrotor flight in
  dense indoor environments.
\newblock In \emph{Robotics Research}, 649--666. Cham: Springer, Switzerland.

\bibitem[{Risb{\o}l and Gustavsen(2018)}]{risbol2018lidar}
Risb{\o}l, O. and Gustavsen, L. (2018).
\newblock Lidar from drones employed for mapping archaeology--potential,
  benefits and challenges.
\newblock \emph{Archaeological Prospection}, 25(4), 329--338.

\bibitem[{Singireddy and Daim(2018)}]{singireddy2018technology}
Singireddy, S.R.R. and Daim, T.U. (2018).
\newblock Technology roadmap: Drone delivery--amazon prime air.
\newblock In \emph{Infrastructure and technology management}, 387--412.
  Springer, Cham, Switzerland.

\bibitem[{Tordesillas et~al.(2021)Tordesillas, Lopez, Everett, and
  How}]{tordesillas2021faster}
Tordesillas, J., Lopez, B.T., Everett, M., and How, J.P. (2021).
\newblock Faster: Fast and safe trajectory planner for navigation in unknown
  environments.
\newblock \emph{IEEE Transactions on Robotics}, 38(2), 922--938.

\bibitem[{Vu et~al.(2021)Vu, Hartl-Nesic, and Kugi}]{vu2021fast}
Vu, M.N., Hartl-Nesic, C., and Kugi, A. (2021).
\newblock Fast swing-up trajectory optimization for a spherical pendulum on a
  7-dof collaborative robot.
\newblock In \emph{Proceedings of the International Conference on Robotics and
  Automation}, 10114--10120.

\bibitem[{Vu et~al.(2022{\natexlab{a}})Vu, Lobe, Beck, Weingartshofer,
  Hartl-Nesic, and Kugi}]{vu2022fast}
Vu, M.N., Lobe, A., Beck, F., Weingartshofer, T., Hartl-Nesic, C., and Kugi, A.
  (2022{\natexlab{a}}).
\newblock Fast trajectory planning and control of a lab-scale 3d gantry crane
  for a moving target in an environment with obstacles.
\newblock \emph{Control Engineering Practice}, 126, 105255.

\bibitem[{Vu et~al.(2022{\natexlab{b}})Vu, Schwegel, Hartl-Nesic, and
  Kugi}]{vu2022sampling}
Vu, M.N., Schwegel, M., Hartl-Nesic, C., and Kugi, A. (2022{\natexlab{b}}).
\newblock {Sampling-Based Trajectory (re) planning for Differentially Flat
  Systems: Application to a 3D Gantry Crane}.
\newblock \emph{arXiv preprint arXiv:2209.05573}.

\bibitem[{Vu et~al.(2020)Vu, Zips, Lobe, Beck, Kemmetm{\"u}ller, and
  Kugi}]{vu2020fast}
Vu, M., Zips, P., Lobe, A., Beck, F., Kemmetm{\"u}ller, W., and Kugi, A.
  (2020).
\newblock Fast motion planning for a laboratory 3{D} gantry crane in the
  presence of obstacles.
\newblock \emph{IFAC-PapersOnLine}, 54(1), 7--12.

\end{thebibliography}
